\documentclass[letterpaper, 10 pt, conference]{ieeeconf}  

\usepackage{amsmath}
\usepackage{graphicx}
\usepackage{hyperref}
\usepackage{color}
\usepackage[style=ieee]{biblatex}

\DeclareSourcemap{
  \maps{
    \map{
      \pertype{article}
      \step[fieldset=url, null]
      \step[fieldset=doi, null]
      \step[fieldset=issn, null]
      \step[fieldset=isbn, null]
      \step[fieldset=note, null]
      \step[fieldset=editor, null]
      \step[fieldset=urldate, null]
      \step[fieldset=file, null]
    }
  }
}
\DeclareSourcemap{
  \maps{
    \map{
      \pertype{inproceedings}
      \step[fieldset=url, null]
      \step[fieldset=doi, null]
      \step[fieldset=issn, null]
      \step[fieldset=isbn, null]
      \step[fieldset=note, null]
      \step[fieldset=editor, null]
      \step[fieldset=urldate, null]
      \step[fieldset=file, null]
    }
  }
}
\DeclareSourcemap{
  \maps{
    \map{
      \pertype{incollection}
      \step[fieldset=url, null]
      \step[fieldset=doi, null]
      \step[fieldset=issn, null]
      \step[fieldset=isbn, null]
      \step[fieldset=note, null]
      \step[fieldset=editor, null]
      \step[fieldset=urldate, null]
      \step[fieldset=file, null]
    }
  }
}
\DeclareSourcemap{
  \maps{
    \map{
      \pertype{misc}
      \step[fieldset=url, null]
      \step[fieldset=doi, null]
      \step[fieldset=issn, null]
      \step[fieldset=isbn, null]
      \step[fieldset=note, null]
      \step[fieldset=editor, null]
      \step[fieldset=urldate, null]
      \step[fieldset=file, null]
    }
  }
}

\IEEEoverridecommandlockouts                              

\title{\LARGE \bf
Reinforcement Learning-Based Model Matching to Reduce the  Sim-Real Gap in COBRA 
}

\author{Adarsh Salagame$^{1\text{\textdagger}}$, Harin Kumar Nallaguntla$^{1\text{\textdagger}}$, Bardia Ardakanian$^{3}$, \\ Eric Sihite$^{2}$, Gunar Schirner$^{1}$, Alireza Ramezani$^{1*}$
\thanks{$^{1}$This author is with the Department of Electrical and Computer Engineering, Northeastern University, Boston MA
        {\tt\small salagame.a, nallaguntla.h, G.Schirner, a.ramezani@northeastern.edu*}}%
\thanks{$^{2}$ This author is with California Institute of Technology, Pasadena CA
		{\tt\small esihite@caltech.edu}}%
\thanks{$^{3}$ This author is with the University of Saskatchewan
            {\tt\small bardia.ardakanian@usask.ca}}
\thanks{\text{\textdagger}These authors have equal contribution to this work.}
\thanks{$*$Indicates the corresponding author.}
}

\begin{document}

\maketitle
\thispagestyle{empty}
\pagestyle{empty}

\begin{abstract}

This paper employs a reinforcement learning-based model identification method aimed at enhancing the accuracy of the dynamics for our snake robot, called COBRA. Leveraging gradient information and iterative optimization, the proposed approach refines the parameters of COBRA's dynamical model such as coefficient of friction and actuator parameters using experimental and simulated data. Experimental validation on the hardware platform demonstrates the efficacy of the proposed approach, highlighting its potential to address the sim-to-real gap in robot implementation.

\end{abstract}

\section{Introduction}
\label{sec:intro}

Snake robots exhibit a multitude of actuated joints, engaging in locomotion patterns involving intricate interactions with the environment \cite{rollinson_gait-based_2013}. These systems present formidable challenges in modeling and control due to the complex interplay of unilateral contact forces, leading to intricate complementarity conditions \cite{studer_numerics_2009}. Traditional approaches have previously tackled these force inclusion issues with promising outcomes \cite{cleach_fast_2023,carius_trajectory_2018}.

Recent advancements in learning methodologies offer expedited solutions, circumventing the need for manually tuned parameter sets and expanding beyond constrained or supervised settings \cite{deshpande_deepcpg_2023,liu_learning_2020,wang_cpg-inspired_2017,gong_feedback_2019,hoeller_anymal_2023}. The models in these studies are further simplified through various assumptions to minimize the number of adjustable parameters. While these methods eloquently showcase how neural modulations can give rise to diverse locomotion strategies \cite{yu_sim--real_2019,qin_sim--real_2019,li_reinforcement_2021,johannink_residual_2019,lee_learning_2020}, there remains limited exploration on extending these models to actively utilize high-dimensional complex observations from onboard sensors to modulate actuators.

\begin{figure}[t]
    \centering
    \includegraphics[width=0.9\linewidth]{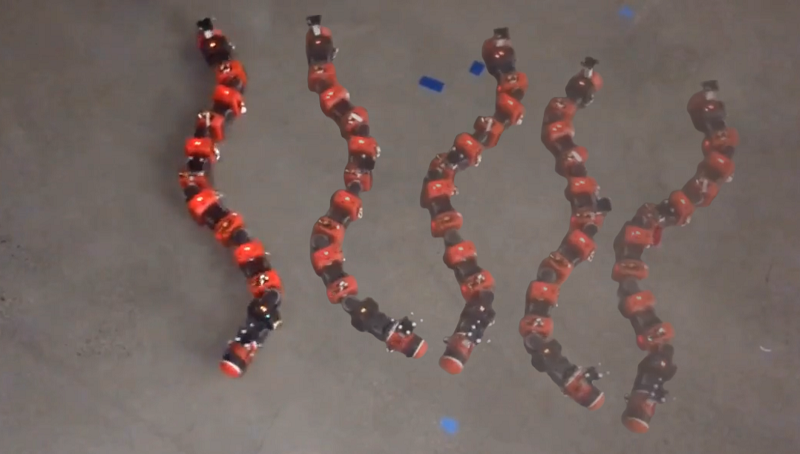}
    \caption{Illustrates COBRA platform while performing sidewinding over flat ground.}
    \label{fig:cover-image}
\end{figure}

This paper addresses the "sim-to-real problem" encountered in the COBRA snake robot \cite{salagame_how_2023, jiang_snake_2023, jiang_hierarchical_2023}, as depicted in Fig.~\ref{fig:cover-image}, where substantial disparities in joint angle trajectories and final head positions emerge between the physical robot and its mathematical model. COBRA is capable of executing programmed 3D interactions with the environment surface.

Driven by the necessity to refine the accuracy of COBRA's locomotion model for effective policy transfer, we employ the Proximal Policy Optimization (PPO) algorithm \cite{schulman_proximal_2017} to determine model parameters, encompassing inertial terms, friction coefficients, and unknown actuator parameters, to synchronize the model's behavior with the hardware platform during predefined locomotion patterns.

\begin{figure}[t]
    \centering
    \includegraphics[width=1\linewidth]{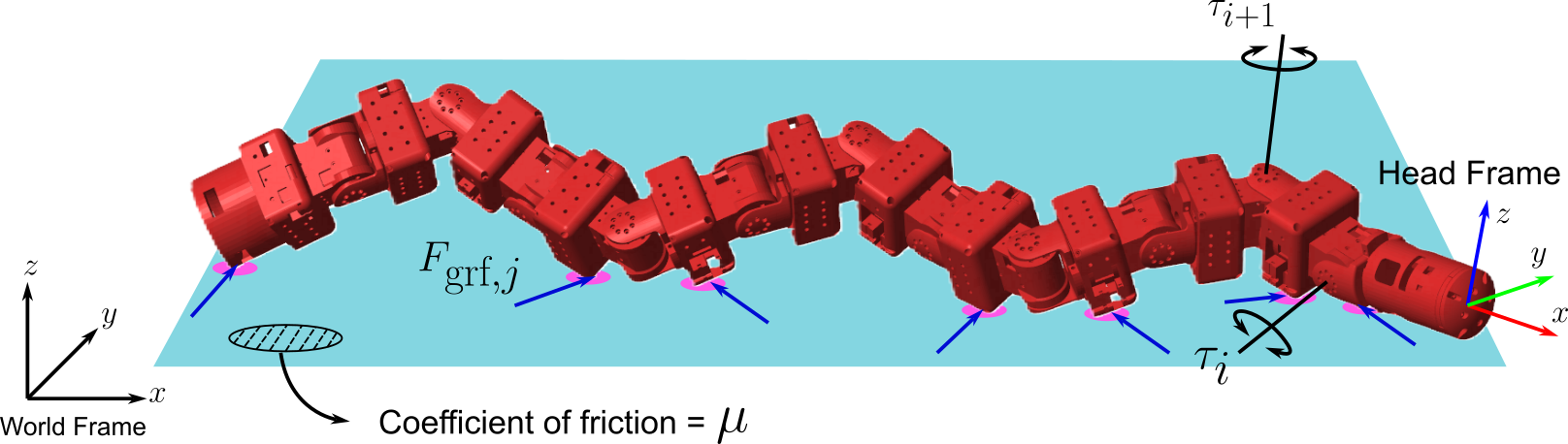}
    \caption{Shows free-body-diagram of COBRA and model parameters unknown to us and found as part of this work.}
    \label{fig:cobra-fbd}
\end{figure}

Our proposed reinforcement learning-based approach aims to bridge the simulation-reality gap, enhancing predictive accuracy and facilitating the transfer of acquired control policies. This research significantly advances snake robot locomotion by addressing the challenges of sim-to-real disparities, thereby contributing to the formulation of robust and adaptable control strategies in complex environments.

\subsection{Quick Overview of COBRA}

COBRA, an acronym for Crater Observing Bio-inspired Rolling Articulator, is a snake robot designed after the intricate locomotion of serpentine creatures. This robot is specifically engineered to navigate and endure rugged terrains and environments, such as craters, where traditional wheeled or legged robots may struggle \cite{meirion-griffith_accessing_2018, marvi_sidewinding_2014, kolvenbach_traversing_2021}. 

Comprising 11 articulated joints, COBRA exhibits flexibility and fluidity in its movements, enabling it to perform a variety of maneuvers, including sidewinding or forming a loop-shape configuration for tumbling. Each module in COBRA emboides a LIPO battery, an actuator and a microprocessor. The modules are connected to each other in a daisy chain. The COBRA robot is equipped with a C++ API that controls a series of Dynamixel servos.

The paper's structure is as follows. We go over the dynamics of Cobra in Section \ref{sec:mdl}, then present the reinforcement learning framework used to identify unknown parameters in the model in Section \ref{sec:method}. Finally, we present results showing the change in performance of the model after training and quantify the improvements made.

\section{COBRA Full-dynamics}
\label{sec:mdl}

The dynamical equations of motion of COBRA is given by
\begin{equation}
    \begin{aligned}
        \left[\begin{array}{cc}
D_H & D_{H a} \\
D_{a H} & D_a
\end{array}\right]\left[\begin{array}{l}
\ddot{q}_H \\
\ddot{q}_a
\end{array}\right]&+\left[\begin{array}{l}
H_H \\
H_a
\end{array}\right]=\left[\begin{array}{l}
0 \\
B_a
\end{array}\right]u+\\&
+
\left[\begin{array}{cc}
J_{H}^\top \\ J_{a}^\top
\end{array}\right]F_{GRF}
    \end{aligned}
\label{eq:full-dynamics}
\end{equation}
\noindent where $D_i$, $H_i$, $B_i$, and $J_i$ are partitioned model \cite{ramezani_atrias_2012,dangol_feedback_2020,ramezani_performance_2013} parameters corresponding to the head 'H' and actuated 'a' joints. $F_{GRF}$ denotes the ground reaction forces. $u$ embodies the joint actuation torques. The model considered for actuators is adopted from M. Spong's book \cite{noauthor_robot_nodate}. The actuator parameters include: transmission inertia, internal damping, and dc motor constant. The reference trajectories used in the actuator models are generated by 
\begin{equation}
\begin{aligned}
    y(t) = A \sin(\omega t + \phi)
\end{aligned}
\label{eq:cpg}
\end{equation}
\noindent where $A$ is the amplitude, $\omega$ is the frequency, and $\phi$ is the phase difference of the signal respectively.

\begin{figure}[t]
    \centering
    \includegraphics[width=0.8\linewidth]{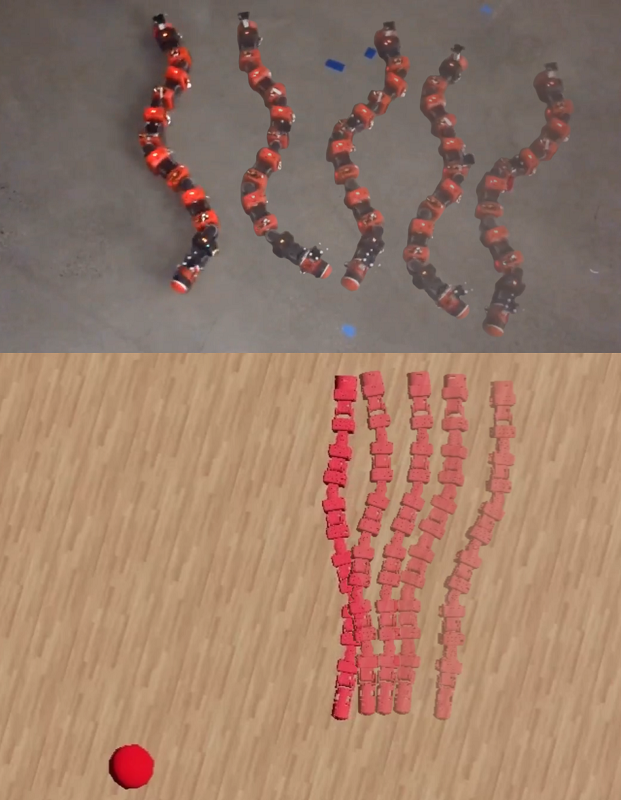}
    \caption{Highlights a significant disparity between the behaviors of COBRA in experiments and simulations. The paper's primary contribution lies in aligning the dynamic behavior of COBRA across both simulation and real-world experiments.}
    \label{fig:sim-to-real-gap-issue}
\end{figure}

\begin{figure}[t]
\begin{center}
    \centering
    \includegraphics[width=1.0\linewidth]{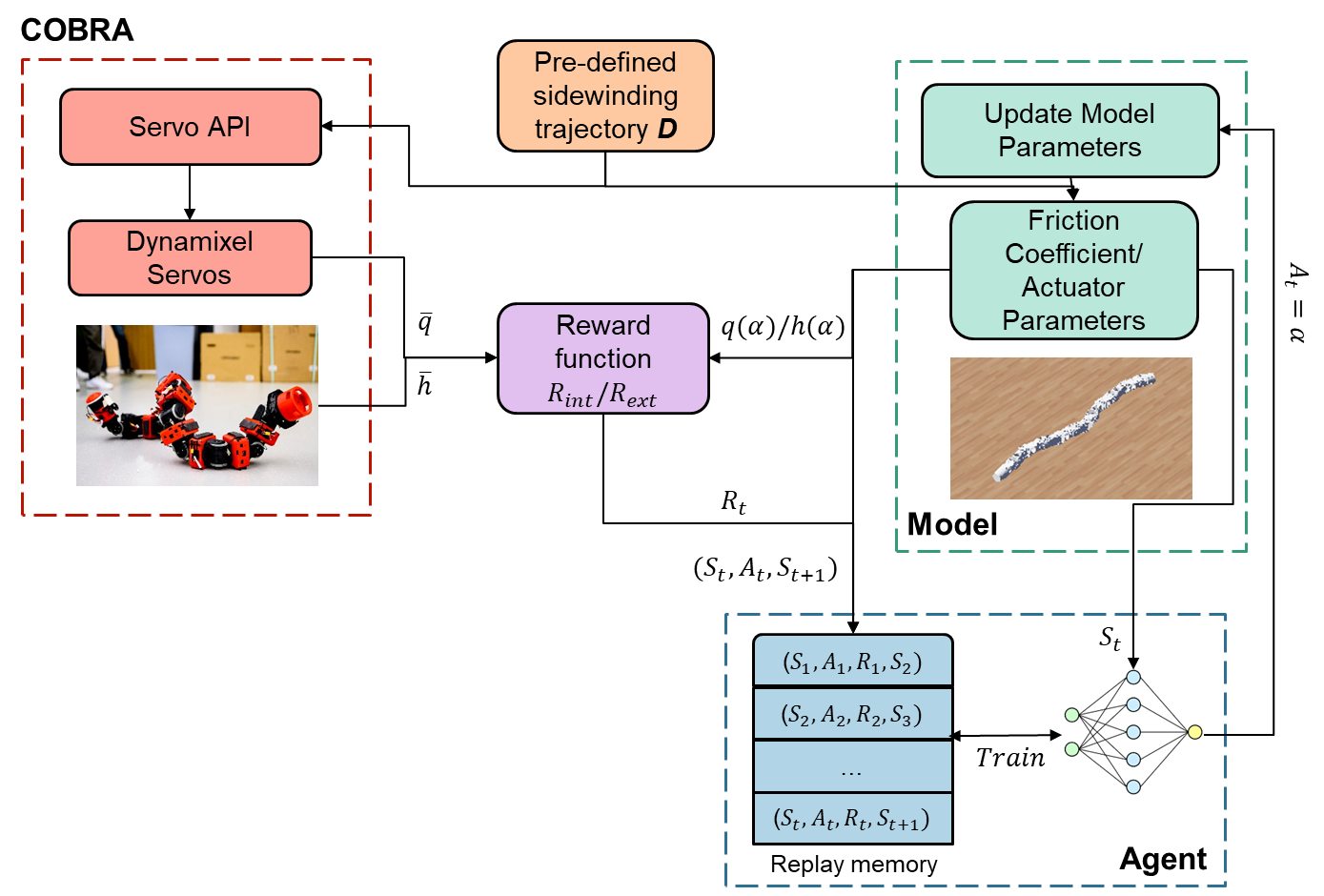}
    \caption{Reinforcement Learning-guided model identification setup used in this work.}
    \label{fig:rl-guided-setup}
\end{center}
\end{figure}

\begin{figure*}[t]
    \centering
    \includegraphics[width=0.8\linewidth]{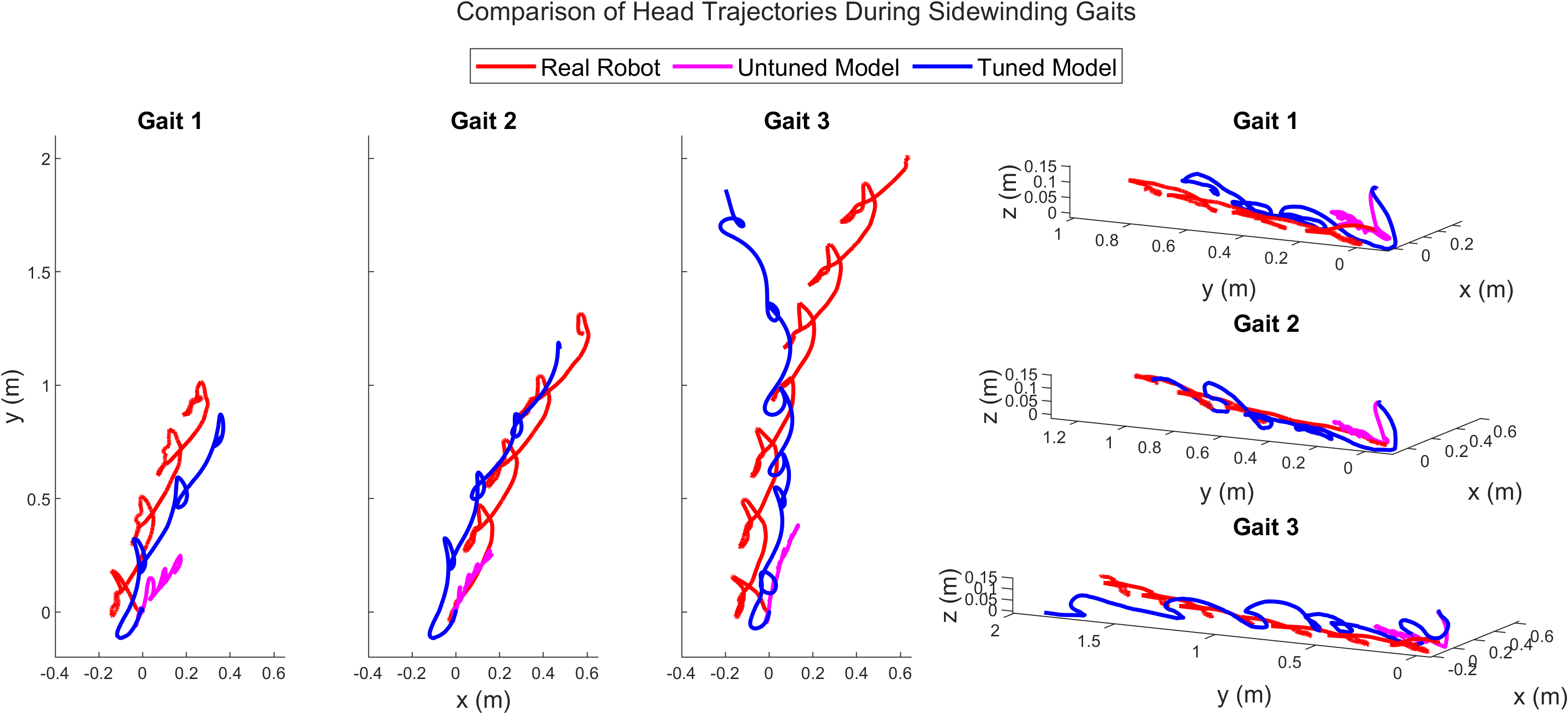}
    \caption{Illustrates a comparison between the head positions in the actual hardware platform (blue), tuned model (orange) and untuned model (red) for a sidewinding trajectory \@ 0.35, 0.5, and 0.65 Hz.}
    \label{fig:trajectory-comparison}
\end{figure*}

The ground model \cite{liang_rough-terrain_2021,dangol_control_2021,salagame_quadrupedal_2023,sihite_efficient_2022-2} used in our simulations is given by 
\begin{equation}
\begin{aligned}
    F_{GRF} &= \begin{cases} \, 0 ~~  \mbox{if } p_{C,z} > 0  \\
     [F_{GRF,x},\, F_{GRF,y},\, F_{GRF,z}]^\top ~~ \mbox{else} \end{cases} \\
\end{aligned}
\label{eq:grf}
\end{equation}
\noindent where $p_{C,i},~~i=x,y,z$ are the $x-y-z$ positions of the contact point; $F_{GRF,i},~~i=x,y,z$ are the $x-y-z$ components of the ground reaction force assuming a contact takes place between the robot and the ground substrate. In Eq.~\ref{eq:grf}, the force terms are given by 
\begin{equation}
     F_{GRF,z} = -k_1 p_{C,z} - k_{2} \dot p_{C,z}, 
    \label{eq:grf-z}
\end{equation}
\begin{equation}
    F_{GRF,i} = - s_{i} F_{GRF,z} \, \mathrm{sgn}(\dot p_{C,i}) - \mu_v \dot p_{C,i} ~~  \mbox{if} ~~i=x, y
    \label{eq:grf-xy}
\end{equation}
\noindent In Eqs.~\ref{eq:grf-z} and \ref{eq:grf-xy}, $k_{1}$ and $k_{2}$ are the spring and damping coefficients of the compliant surface model. In Eq.~\ref{eq:grf-xy}, the term $s_i$ is given by 
\begin{equation}
    s_{i} = \Big(\mu_c - (\mu_c - \mu_s) \mathrm{exp} \left(-|\dot p_{C,i}|^2/v_s^2  \right) \Big)
    \label{eq:grf-s}
\end{equation}
\noindent where $\mu_c$, $\mu_s$, and $\mu_v$ are the Coulomb, static, and viscous friction coefficients; and, $v_s > 0$ is the Stribeck velocity. Specifically speaking, from Eqs.~\ref{eq:full-dynamics} and \ref{eq:grf}, the following parameters are unknown to us: actuator models parameters and Stribeck terms.

In Section.~\ref{sec:method}, we will use RL to obtain these unknown parameters. The unknown model parameters are encoded within the simulator as part of sim-to-real gap research. 

\subsection{Selection of Simulator}
To provide a good starting point for the model matching, the right simulator must be selected. Simulators such as Matlab Simscape, Drake and MuJoCo are commonly used for simulating dynamics, and others such as Gazebo are more suited for high level perception and navigation tasks. In this work, Webots is chosen for its robust physics simulation capabilities and adaptability to multi-degree-of-freedom robot locomotion complexities, as well as the ability to easily add several sensing modalities for training a closed loop locomotion policy in the future. Developed by Cyberbotics, Webots offers a realistic and dynamic simulation environment, accurately modeling interactions between the COBRA snake robot and its surroundings. Its physics engine enables precise evaluation of robot behavior while executing predefined trajectories. Webots also has a user-friendly interface and extensive documentation, and is a popular choice in the RL community for its ability to change model parameters programmatically with ease, enabling encoding unknown model parameters for Learning applications.

\section{Reinforcement Learning-guided Identification of Actuator and Stribeck Parameters}
\label{sec:method}

\subsection{Urgency}

Figure.~\ref{fig:sim-to-real-gap-issue} illustrates the behavior gap between the hardware platform and mathematical model. The sim-to-real problem in COBRA locomotion manifests as variations in final head positions between the simulated robot and its physical counterpart when exactly the same joint motions are commanded in both models. These disparities hinder the seamless transfer of control policies developed in simulation to the real robot, limiting the effectiveness of learned behaviors in practical applications. Hence, addressing the sim-to-real issue is crucial for ensuring the reliability and robustness of COBRA when deployed in diverse and dynamic environments.

\subsection{Approach}

Our methodology provides a structured approach to bridging the sim-to-real gap for the COBRA robotic system. We combine physical hardware, a virtual simulation environment, and a reinforcement learning framework to iteratively refine the fidelity of the simulation, thereby ensuring that it mirrors the real-world performance of the robot as closely as possible.

The simulation of the COBRA robot is conducted within Webots, which incorporates an ODE physics engine to reproduce the dynamics of robotic systems. The physics engine takes into consideration the mechanical properties, such as mass and friction, as well as the actuator models to simulate the robot's interaction with its environment.

\subsection{Performance Comparison Based on Underactuated and Actuted Dynamics from Eq.~\ref{eq:full-dynamics}}

A sidewinding trajectory command $q_r$ is employed to guide both the real COBRA robot and its virtual model through identical movements. The performance analysis occurs in two separate domains as suggested by the partitioned model in Eq.~\ref{eq:full-dynamics}:

\begin{itemize}
    \item \textit{Underactuated dynamics:} The imposition of $q_r$ leads to the head translations $\ddot q_H$, which are underactuated. Comparing the head's position and orientation provides a clear performance metric for discerning contact force discrepancies (friction coefficients) between the physical robot and the simulation while completely excluding actuator dynamics.
    
    \item \textit{Actuated dynamics:} In addition to matching passive dynamics models, we also address actuated dynamics, specifically the joint angle trajectories $\ddot q_a$ in Eq.~\ref{eq:full-dynamics}. Aligning actuator models is crucial for evaluating the precision of the robot's movements. This comparison can be accomplished by calculating the norm distance of output joint motions between the physical robot and the simulated model.
    
\end{itemize}

\subsection{Reinforcement Learning Framework}

The PPO algorithm is used to iteratively refine the model parameters. This is done with the aim of minimizing the observed discrepancies in the underactuated and actuated dynamics between the model and the actual robot. Below, we explain the state, action, and reward function defined for the PPO algorithm.

\subsubsection{State, Action, and Reward Definition}

The state $S_t$ encapsulates the actuator parameters for internal tuning and Steibek terms for external tuning. The actions $A_t$ are the modifications applied to these simulation parameters.

The reward function $R$ is given by
\begin{equation}
\begin{aligned}
    R_{\text{external}} = - \sqrt{\left( (x_{\text{des}} - x_{\text{actual}})^2 + (y_{\text{des}} - y_{\text{actual}})^2 \right)} 
\end{aligned}
\label{eq:external reward}
\end{equation}
where $x_{\text{des}}$ and $y_{\text{des}}$ denote desired x- and y-positions of the head module, respectively. $x_{\text{actual}}$ and $y_{\text{actual}}$ are the OptiTrack data. And, $R_{\text{internal}}$ is given by 
\begin{equation}
\begin{aligned}
    R_{\text{internal}} = &- (\phi_{\text{des}} - \phi_{\text{actual}})^2 - (\omega_{\text{des}} - \omega_{\text{actual}})^2  \\
    &- (A_{\text{des}} - A_{\text{actual}})^2
\end{aligned}
\label{eq:internal reward}
\end{equation}
where $\phi$, $\omega$, and $A$ are CPG variables. These reward functions are crafted to penalize the deviation in both passive dynamics and actuated dynamics. They are defined as the sum of the L$^2$-norm of the joint angle trajectory differences and the L$^2$-norm of the final head position differences.

\subsubsection{Policy Updates}

The policy search recruited here operates in a cycle of simulation runs and updates. During simulation runs, the algorithms collects data on state transitions and rewards, which are aggregated into sequences $(S_t, A_t, R_t, S_{t+1})$ and stored in a replay memory. The training phase involves updating the policy network with this data, where PPO adjusts the policy in a manner that maximizes the cumulative reward while maintaining a degree of similarity to the previous policy, using a mechanism known as clipping to avoid drastic policy changes \cite{schulman_proximal_2017}.

Through these iterative training cycles, PPO tunes the model parameters to align the model performance with the actual COBRA robot, following a predefined sidewinding trajectory. The process continues until a convergence is reached. The following is a line-by-line breakdown of the search approach as reported in \cite{schulman_proximal_2017}:

\textit{a) Input:} The algorithm starts with initial policy parameters $\theta_0$ and initial value function parameters $\phi_0$. These parameters are what the algorithm will learn to adjust as it interacts with the environment to improve its policy.

\textit{b) Collect Trajectories:} A set of trajectories $D_k$ is collected by running the current policy $\pi(\theta_k)$ in the environment. A trajectory is a sequence of states, actions, and rewards experienced by the agent.

\textit{c) Compute rewards-to-go:} For each time step $t$, compute the total expected rewards from that time step until the end of the trajectory, denoted as $R_t$. This is used to estimate how good it is to be in a particular state.

\textit{d) Compute advantage estimates:} The advantage $A_t$ indicates how much better or worse an action is compared to the average action in a given state. This is calculated using the value function $V_{\phi}$ and the rewards-to-go.

\textit{e) Update the Policy:} First, we calculate the probability ratio, which is a fraction where the numerator is the probability of taking action $a_t$ in state $s_t$ under the new policy $\pi_{\theta}$, and the denominator is the probability of taking action $a_t$ in state $s_t$ under the old policy $\pi_{\theta_{\text{old}}}$:
\[ \text{ratio}(\theta) = \frac{\pi_{\theta}(a_t | s_t)}{\pi_{\theta_{\text{old}}}(a_t | s_t)} \]
This ratio measures how the new policy differs from the old policy in terms of the likelihood of taking the same actions.

We then clip this ratio to be within a range of $[1 - \epsilon, 1 + \epsilon]$, where $\epsilon$ is a hyperparameter typically set to a small value like 0.1 or 0.2. This clipping limits the amount by which the new policy can differ from the old one, regardless of the advantage estimate:
\[ \text{clipped}(\theta) = \text{clip}(\text{ratio}(\theta), 1 - \epsilon, 1 + \epsilon) \]
The objective function incorporates the clipped ratio and the advantage function $A_t$. It takes the minimum of the unclipped and clipped objectives, ensuring that the final objective doesn't take too large of a step (hence, "clipping"):
\[ L(\theta) = \min \left( \text{ratio}(\theta) A_t, \ \text{clipped}(\theta) A_t \right) \]
The expectation of this objective function over all timesteps and all trajectories is what the algorithm seeks to maximize. This expectation is approximated by averaging over a finite batch of timesteps and trajectories:
\[ \theta_{k+1} = \arg\max_{\theta} \left( \frac{1}{\left|D_k\right| T} \sum_{\tau \in D_k} \sum_{t=0}^{T} L(\theta) \right) \]
Finally, the parameters of the policy $\theta$ are updated using stochastic gradient ascent. This means we compute gradients of the objective function with respect to the policy parameters and adjust the parameters in the direction that increases the objective:
\[ \theta_{\text{new}} \leftarrow \theta_{\text{old}} + \alpha \nabla_{\theta} \left( \frac{1}{\left|D_k\right| T} \sum_{\tau \in D_k} \sum_{t=0}^{T} L(\theta) \right) \]
where $\alpha$ is the learning rate.

The Adam optimizer is often used for this step because it adapts the learning rate for each parameter, helps in converging faster, and is more robust to the choice of hyperparameters.

By maximizing this objective, PPO-Clip seeks to improve the policy by making sure that the actions that would increase the expected return are taken more frequently, while ensuring that the policy does not change too drastically, which could lead to poor performance due to overfitting to the current batch of data.

\section{RESULTS}

\begin{figure}[t]
    \centering
    \includegraphics[width=0.7\linewidth]{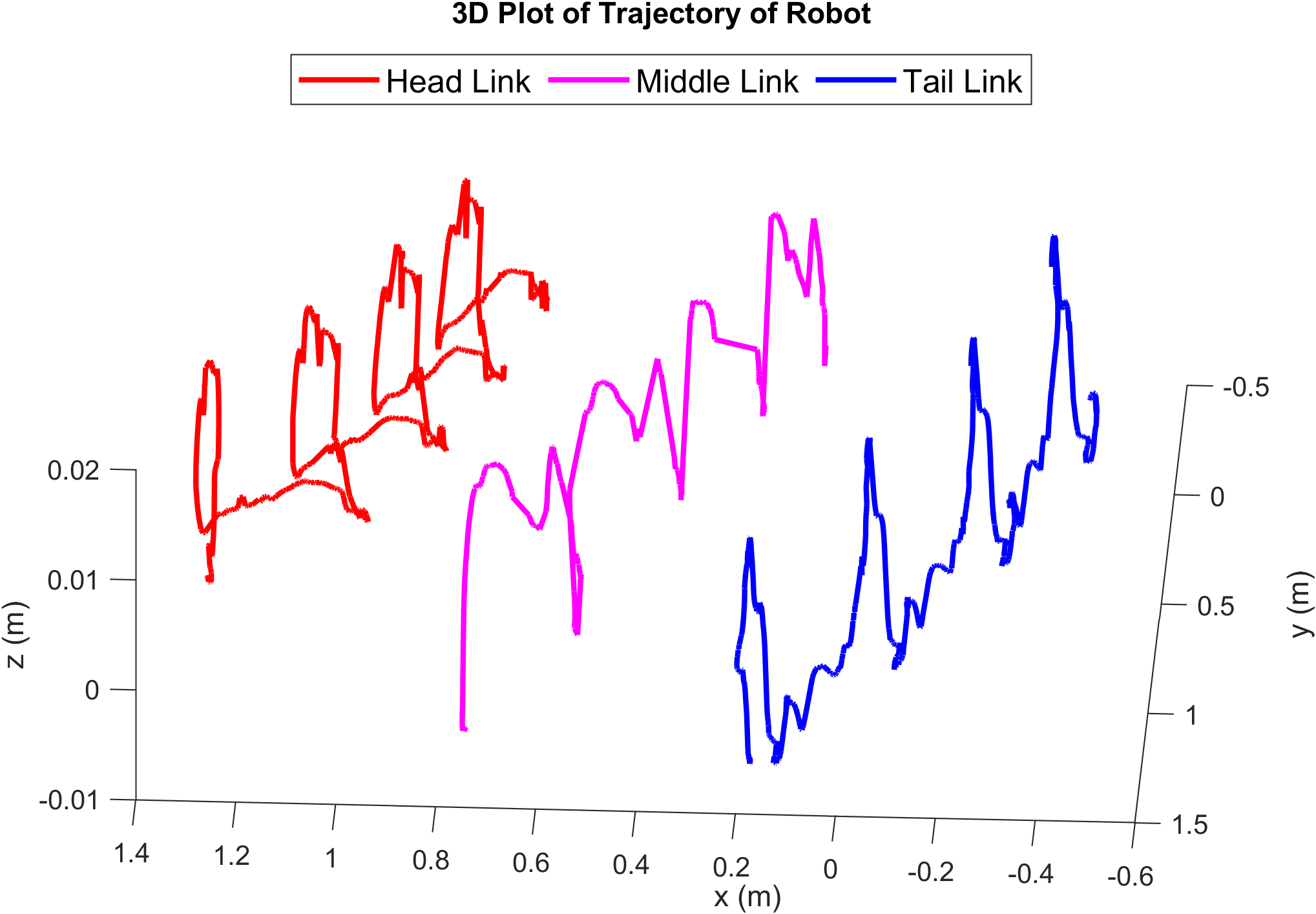}
    \caption{Illustrates body positions (head module, link 6, and tail module) during slithering towards the desired position on a flat ground collected from the actual hardware platform using an OptiTrack system.}
    \label{fig:optitrack-trajectory}
\end{figure}

\begin{figure}[t]
    \centering
    \includegraphics[width=0.8\linewidth]{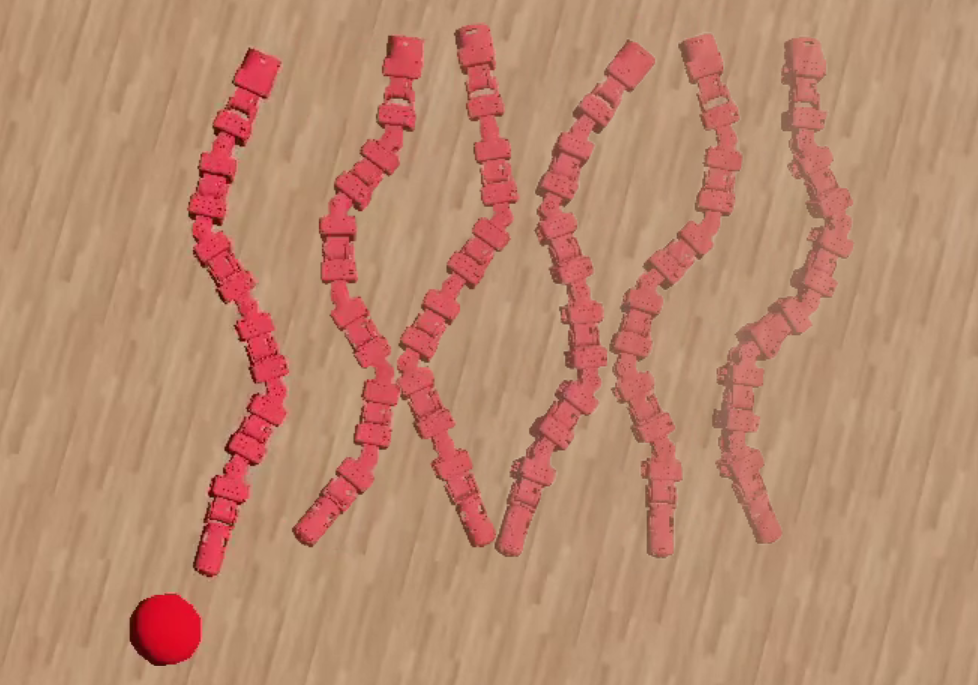}
    \caption{Show COBRA in Webots simulator with tuned parameters performing the sidewinding motion. The red ball shows the location the actual robot achieved when similar joint trajectories were applied.}
    \label{fig:tuned_snapshots}
\end{figure}

\begin{figure}[t]
    \centering
    \includegraphics[width=0.8\linewidth]{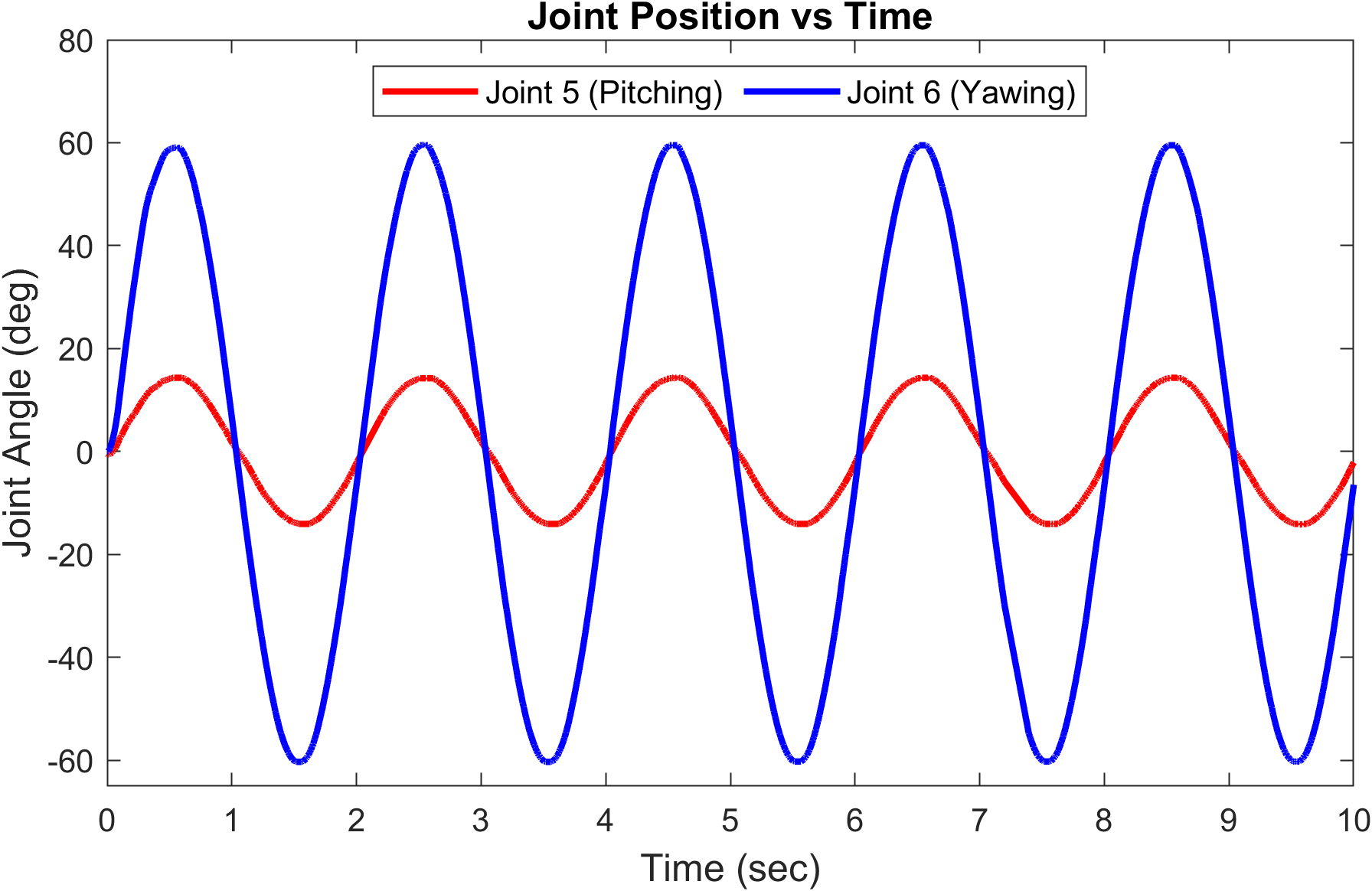}
    \caption{Shows joint angles (e.g., joints 5 and 6) collected from the hardware platform.}
    \label{fig:joint_traj}
\end{figure}

\begin{figure}[t]
    \centering
    \includegraphics[width=0.8\linewidth]{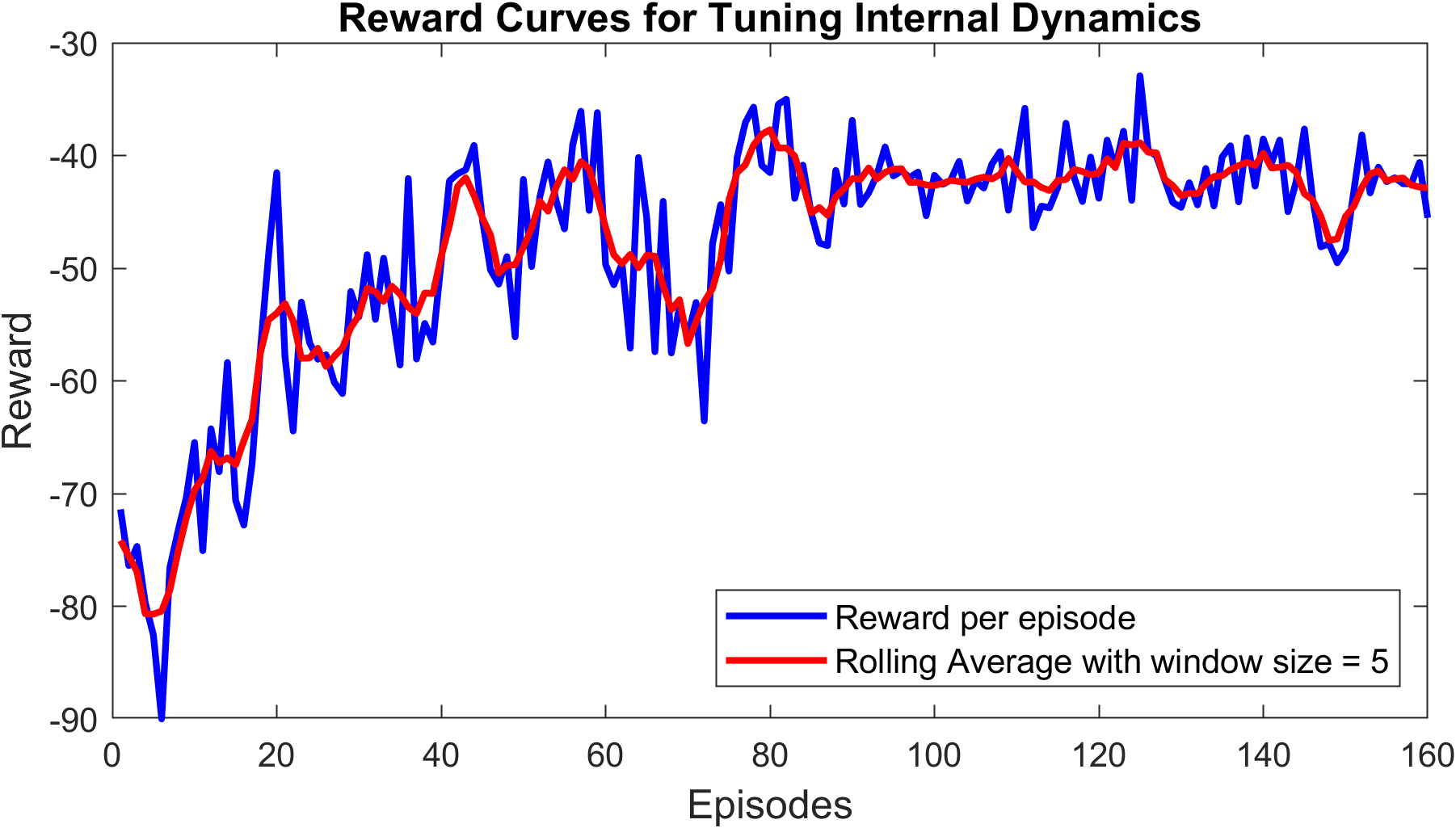}
    \caption{Shows the training reward curve for tuning model parameters.}
    \label{fig:reward}
\end{figure}

\begin{figure}[t]
    \centering
    \includegraphics[width=0.7\linewidth]{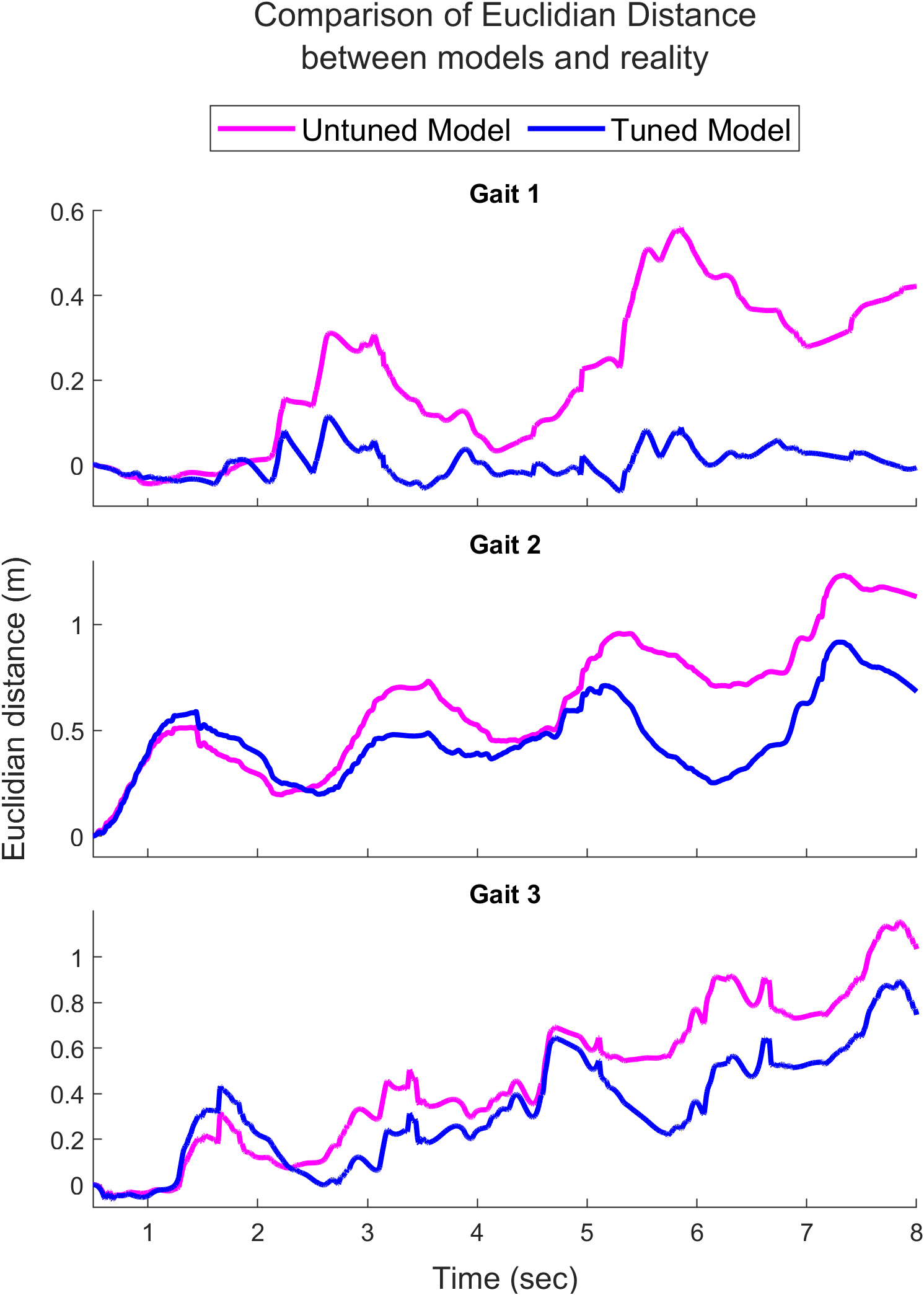}
    \caption{Comparison of the Euclidean distance (error metric)  between the actual platform's head position captured by OptiTrack and tuned/untuned models for sidewinding @ 0.35, 0.5, and 0.65 Hz.}
    \label{fig:euclidean}
\end{figure}

\begin{figure}[t]
    \centering
    \includegraphics[width=1\linewidth]{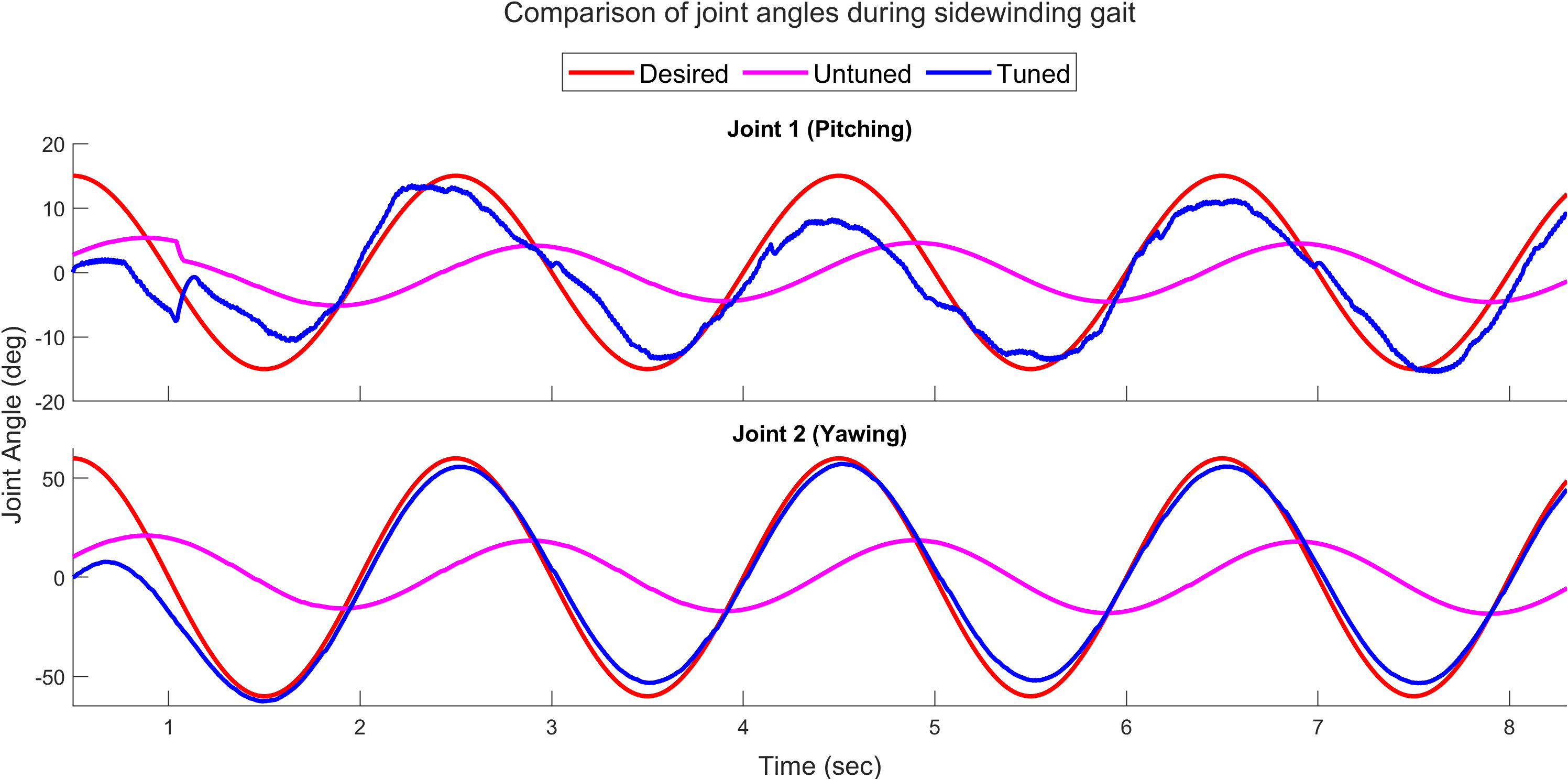}
    \caption{Shows a comparison between the actuator joint responses from the actual hardware platform (red), the tuned (blue) and untuned (magenta) models for a sidewinding gait at @ 0.5 Hz.}
    \label{fig:joint-traj-comparison}
\end{figure}

Experiments were performed on the real robot and motion capture data for the precise trajectory of the head link, the tail link, and the middle link was recorded. This is illustrated in Fig. \ref{fig:optitrack-trajectory}. In addition, joint trajectory data for all eleven joints were recorded. Odd-numbered joints (1,3,5,...,11 from the head) execute a pitching motion and even-numbered joints (2,4,...,10 from the head) execute a yawing motion. A sidewinding gait created by supplying a sinusoidal input signal to each of the joints is used as a standard input with varying parameters to change the behavior of the robot. The input signal is given by a sine-wave according to Eq. \ref{eq:cpg} with some phase difference for each joint, as follows:
$$
A_\text{pitching} = 14^\circ,~A_\text{yawing} = 60^\circ
$$
$$
\phi = \frac{\pi}{2}[0, 0, 1, 1, 2, 2, 3, 3, 0, 0, 1]
$$
The frequency for this sine-wave input was varied for diversity of data, creating three gaits with frequencies $0.35$ Hz, $0.5$ Hz, $0.65$ Hz respectively. The recorded joint trajectories for two joints performing the $0.5$ Hz frequency input is shown in Fig. \ref{fig:joint_traj}. The actuators on the robot almost perfectly track the desired input signal. A total of 25 experiments were conducted, recording the robot moving from various initial conditions for the given set of three gaits. The robot performance was highly repeatable and one representative sample was kept for validation while the remaining was used for training of the simulation model.

\begin{figure}[t]
    \centering
    \includegraphics[width=0.8\linewidth]{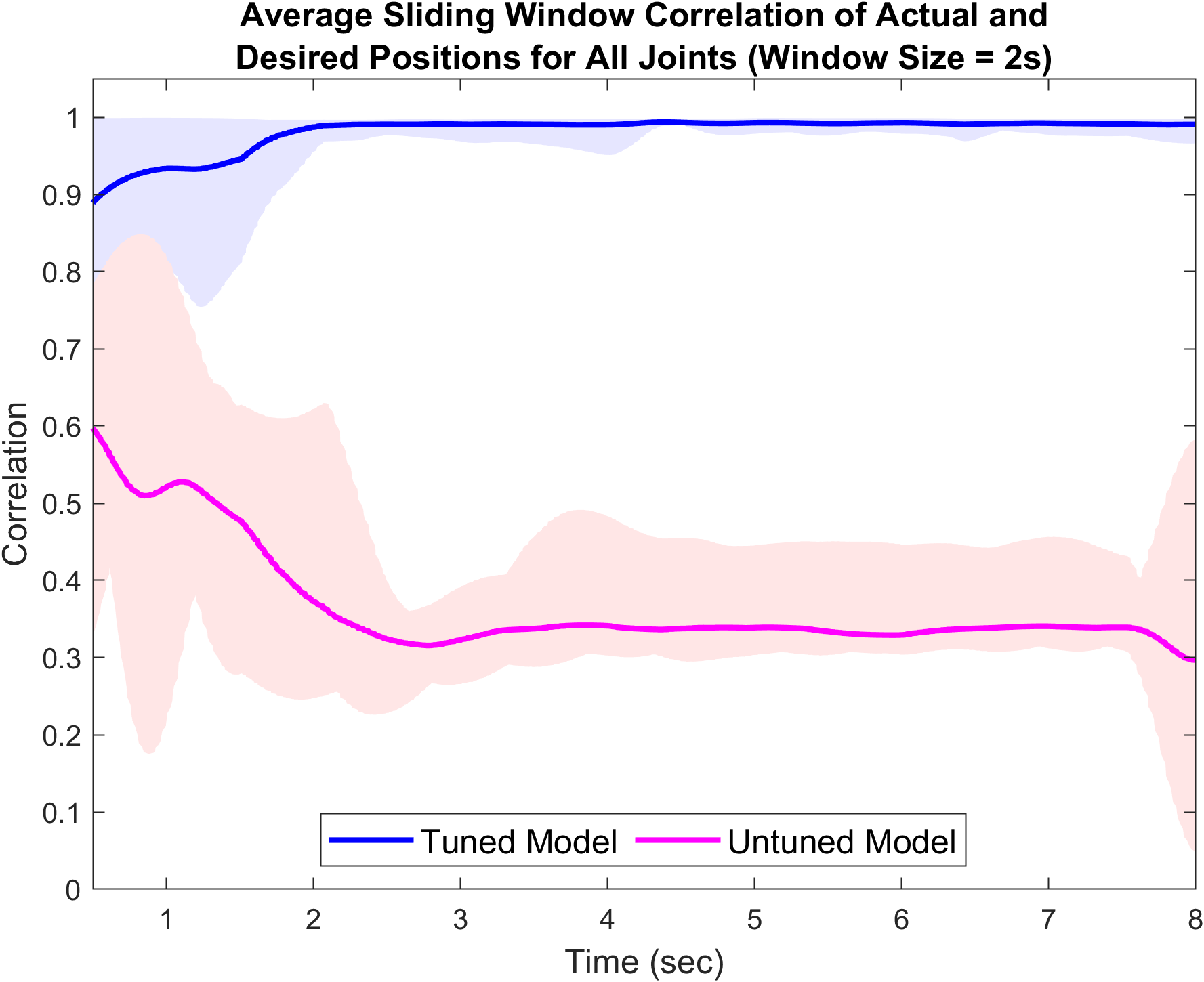}
    \caption{Shows a comparison of average sliding window correlation (obtained by taking average of all the 11 correlation curves displayed in the above plot) of yawing joint positions for tuned and default simulators with desired signal (for the sidewinding gait @ 0.5 Hz). The model matching improved the actuator model in the the simulator as seen from the improved average correlation factor between desired and actual joint positions. Shaded regions represent the range between min and max correlation across all joints.}
    \label{fig:correlation}
\end{figure}

Each of the three desired gaits was then executed in simulation, matching the initial conditions of the robot to reality for all training data. Fig. \ref{fig:sim-to-real-gap-issue} presents snapshots that illustrate the difference in the behavior of the robot in simulation as compared to reality before tuning. The model was then tuned by training using the framework presented in Fig. \ref{fig:rl-guided-setup}. Fig. \ref{fig:reward} shows the reward curve from tuning the internal dynamics of the robot model. Fig. \ref{fig:tuned_snapshots} presents snapshots of the same experiment in simulation after tuning. The trajectory of the head module is compared in Fig. \ref{fig:trajectory-comparison}, showing significant improvements with the simulation matching the trajectory of the real robot. The Euclidean distance between the reference trajectory from the real robot and the trajectory from the simulation is plotted in Fig. \ref{fig:euclidean}, showing that for all three gaits, the error for the tuned model is lower. From this figure, it may be observed that as the movement of the robot becomes more aggressive (by increasing the frequency of the input sine wave), the tuned model reduces in performance. This is expected to some degree but can be mitigated by increasing the diversity of training data to improve the quality of the model. The performance is further inspected at the joint level by comparing the joint trajectories of the simulation model with the desired trajectories. Fig. \ref{fig:joint-traj-comparison} shows the signal for two representative joints, with the tuned model showing significantly closer tracking of the desired signal. Fig. \ref{fig:correlation} tracks the overall performance of all joints using sliding window correlation. 

\section{Conclusion}

Despite limited training data, the framework has significantly improved the ability of the simulation to predict the behaviour of the robot. However, due to the limited training data, the model shows a dropoff in accuracy when attempting to generalize to more environments and types of inputs as observed earlier. This may be solved by increasing the diversity of training data. However, expanding the training set using data from real hardware is expensive both in terms of wear and tear to the robot and the time required for operation. This is circularly related to the primary motivation behind the presented work on model matching and underscores the importance of these results. With the tuned model, a locomotion policy trained in simulation has a higher chance of success when transferred to the real robot, reducing the sim-to-real gap. Demonstrating a successful sim-to-real transfer of locomotion policy will be presented in future work.

\printbibliography

\end{document}